\documentclass[review]{elsarticle}

\usepackage{lineno,hyperref}
\modulolinenumbers[5]

\usepackage{amsmath}
\usepackage{amsfonts} 
\usepackage{mathptmx}
\usepackage{amssymb}
\usepackage[utf8]{inputenc}
\usepackage{siunitx}
\usepackage{subcaption}
\usepackage{multirow, makecell}
\usepackage{xcolor}
\usepackage{todonotes}
\setuptodonotes{size=\small}
\usepackage{booktabs}

\journal{Big Data Research}

\begin{document}

\begin{frontmatter}

\title{Cardiac Cohort Classification based on Morphologic and Hemodynamic Parameters extracted from 4D PC-MRI Data}


\author[kmd]{Uli~Niemann\corref{mycorrespondingauthor}}
\cortext[mycorrespondingauthor]{Corresponding author}

\author[vis]{Atrayee~Neog}

\author[vis]{Benjamin~Behrendt}
\author[jena]{Kai~Lawonn}
\author[leipzig]{Matthias~Gutberlet}
\author[kmd]{Myra~Spiliopoulou}
\author[vis]{Bernhard~Preim}
\author[vis,jena]{Monique~Meuschke}

\address[kmd]{Institute of Technical and Business Information Systems, University of Magdeburg,  Germany}
\address[vis]{Department of Simulation and Graphics, University of Magdeburg, Germany}
\address[jena]{Institute of Computer Science, University of Jena, Germany}
\address[leipzig]{Heart Centre, University of Leipzig, Germany}

\begin{abstract}


An accurate assessment of the cardiovascular system and prediction 
of cardiovascular diseases (CVDs) are crucial. 
Measured cardiac blood flow data provide insights about patient-specific hemodynamics, 
where many specialized techniques have been developed for the visual 
exploration of such data sets to better understand the influence of morphological 
and hemodynamic conditions on CVDs. 
However, there is a lack of machine learning approaches techniques that allow a
feature-based classification of heart-healthy people and patients with CVDs.

In this work, we investigate the potential of morphological and hemodynamic 
characteristics, extracted from measured blood flow data in the aorta, for the 
classification of heart-healthy volunteers and patients with bicuspid aortic valve (BAV). 
Furthermore, we research if there are characteristic features to classify male 
and female as well as older heart-healthy volunteers and BAV patients. 
We propose a data analysis pipeline for the classification of the cardiac status, encompassing feature selection, model training and hyperparameter tuning.
In our experiments, we use several feature selection methods and classification 
algorithms to train separate models for the healthy subgroups and BAV patients.
We report on classification performance and investigate the predictive power of 
morphological and hemodynamic features with regard to the classification of the 
defined groups. 
Finally, we identify the key features for the best models.

\end{abstract}


\end{frontmatter}


\section{Introduction}
\label{sec:intro}
Worldwide, most people die of cardiovascular diseases (CVDs)~\cite{mendis2011}. 
Therefore, an accurate assessment of the cardiovascular system and prediction 
of CVDs are crucial. 
Various factors influence the formation and progression of CVDs. 
Besides abnormal vascular morphology and genetic factors, there appear to be 
relationships between hemodynamics and cardiac pathologies~\cite{franccois2012,geiger2012,lorenz2014,McNally.2017}.

To investigate these relationships, patient-specific flow information is required. 
In clinical routine, two-dimensional phase-contrast magnetic resonance imaging 
flow measurements (2D PC-MRI) are used for the diagnosis and follow-up of 
cardiovascular diseases. 
They allow the non-invasive quantification of flow volumes, amount of backward flow and peak flow 
velocities and thus an assessment of the severity of heart valve defects and the 
determination of general heart function. 
However, 2D PC-MRI has serious limitations. 
The analysis is limited to the selected single layer and one flow direction. 
Adjacent vessel sections or flow components outside the main flow direction 
cannot be assessed. 
Correct angulation of the measuring plane to the vessel course is essential, as 
angle errors lead to incorrect flow values. 
The flow in the heart vessels, such as the aorta, is a highly complex, time-varying 
3D structure that cannot be completely represented by a single measuring plane and a 
main flow vector. 
In addition, no information about turbulence is obtained. 
This is critical because quantification in the vicinity of turbulence is 
unreliable~\cite{Koehler_2015}.

Time-dependent flow data in a 3D volume can be acquired non-invasively using 
four-dimensional phase-contrast magnetic resonance imaging (4D PC-MRI)~\cite{Markl2012}.  
The flow information can be evaluated in any vessel section with freely 
modifiable measurement planes. 
Thus, 4D PC-MRI has great potential to improve the diagnosis, follow-up and 
treatment decisions of CVDs~\cite{Suwa2020}. 

Several groups developed visual exploration techniques for single 
4D PC-MRI data sets~\cite{vanPelt2010, Heiberg2012,Koehler2019,Semaan2014, Wehrum2014} 
involving different flow attributes and solutions for conveying the temporal 
data behavior. 
However, to develop guidelines how to interpret 4D PC-MRI data requires the analysis
of cohorts instead of single data sets. 
Therefore, we study to what extent disease status can be predicted with supervised 
classification models learned on morphological and hemodynamic attributes, which 
were extracted from 4D PC-MRI data of heart-healthy volunteers (HHV) and patients with 
bicuspid aortic valve (BAV). 
To support physicians in diagnosing and assessing the severity of BAV defects 
and to better understand the variety of physiological hemodynamic and morphological 
characteristics, we have formulated three research questions that we aim to answer with 
our classification procedure:  
(i) Do heart-healthy volunteers and BAV patients differ regarding the extracted features~\cite{Ebel2020}?
(ii) Are there predictive features that can separate between older heart-healthy volunteers and BAV patients? 
This question is motivated by the challenge in clinical practice to distinguish between non-pathological decline in cardiac function with increasing age~\cite{Lernfeit1991} (\emph{healthy aging}) and the onset of early stages of BAV, solely based on flow velocity. 
(iii) Do female and male heart-healthy volunteers differ with respect to morphological and hemodynamic attributes? Since women generally have a smaller heart, a smaller blood volume and a faster heartbeat than men~\cite{Legato2004}, the identification of further differences related to aortic blood flow might suggest the necessity to stratify diagnostic procedures or treatment pathways by gender.
We train separate models for the healthy subgroups and BAV patients based on the 
combination of several feature selection methods and learning algorithms. 
Model training is combined with a feature selection step to identify correlations 
between the multitude of flow attributes and to reduce them to a meaningful subset.  
Finally, we study and compare the informativeness of each feature towards the 
outcomes.
Furthermore, the determination of standard values and follow-ups for flow 
attributes is supported. 
\section{Related Work}
\label{sec:relwork}

There is extensive research on machine learning (ML) in cardiology, for example to support computer-aided diagnosis and prediction of heart diseases~\cite{AlAref2019}. 
Dinh et al.~\cite{Dinh2019} created a classification model ensemble from 
different base classifiers in the context of cardiovascular disease diagnosis.
Patient characteristics such as age, blood pressure, body weight and chest pain 
were found to be most predictive. 
Similarly, Miao et al.~\cite{Miao2016} trained an AdaBoost~\cite{Freund:AdaBoost1997} 
ensemble classifier for coronary heart disease diagnosis using data gathered from 
four different medical institutions. 
Wojnarski et al.~\cite{Wojnarski2018} identified three distinct phenotypes of BAV patients based on vascular morphology via data clustering. 
While acknowledging the high potential of ML, Russak et al.~\cite{Russak2020} observed that case studies of a successful integration of these technologies into clinical practice are still sparse. 
They further suggest guidelines to increase acceptance and ultimately to enable clinical adoption of ML techniques in cardiology, including a project development driven by unsolved clinical questions rather than by the available data only, incorporating experts from multiple disciplines, using complete patient data gathered in a standardized format, improving interpretability of complex ML models, and training clinicians on ML fundamentals to increase familiarity with these concepts. 

Recently, deep learning has become increasingly popular also for cardiovascular applications~\cite{Martin2020, Fries2019, Miao2018, Arabasadi:DL2017, Kim2017}. 
In comparison to classic ML algorithms, deep neural networks often exhibit superior 
accuracy and are able to work on raw image data, thus reducing the necessity to 
carefully extract features using domain knowledge which is often too time-consuming 
or expensive. 
However, the ``black-box'' nature of deep neural networks is rather unsuited for 
inference, which is crucial especially in medical applications. 
Thus, there are studies that create more parsimonious models using extracted 
interpretable parameters to better understand characteristics of anatomical 
structures which are predictive for the outcome. 
Niemann et al.~\cite{Niemann_2018} studied the potential of 22 morphological 
parameters extracted from angiographic images for a data-driven rupture risk 
classification in intracranial aneurysms. 
Post-hoc interpretation steps revealed the most predictive features, which 
included a dome point angle, ellipticity index and aneurysm volume. 
In addition to morphology, Detmer et al.~\cite{Detmer:Neuroradiology2019} investigated 
also hemodynamic parameters and patient characteristics for rupture risk assessment 
and found that ruptured aneurysms exhibit larger and more complex flow characteristics. 

Often, the number of extracted features is large, which often leads to unnecessarily 
complex models. 
Selecting a subset of features can help to reduce computational complexity, 
increase model performance and enhance model interpretability~\cite{Wah2018, Latha2019, Parthiban2008}. 
For example, Latha et al.~\cite{Latha2019} combined ensemble classifiers with a 
preceding feature selection step which improved model accuracy of heart disease prediction.
\section{Medical Background}
\label{sec:dataac}
\begin{figure}[t]
   \centering
   \includegraphics[width=0.5\linewidth]{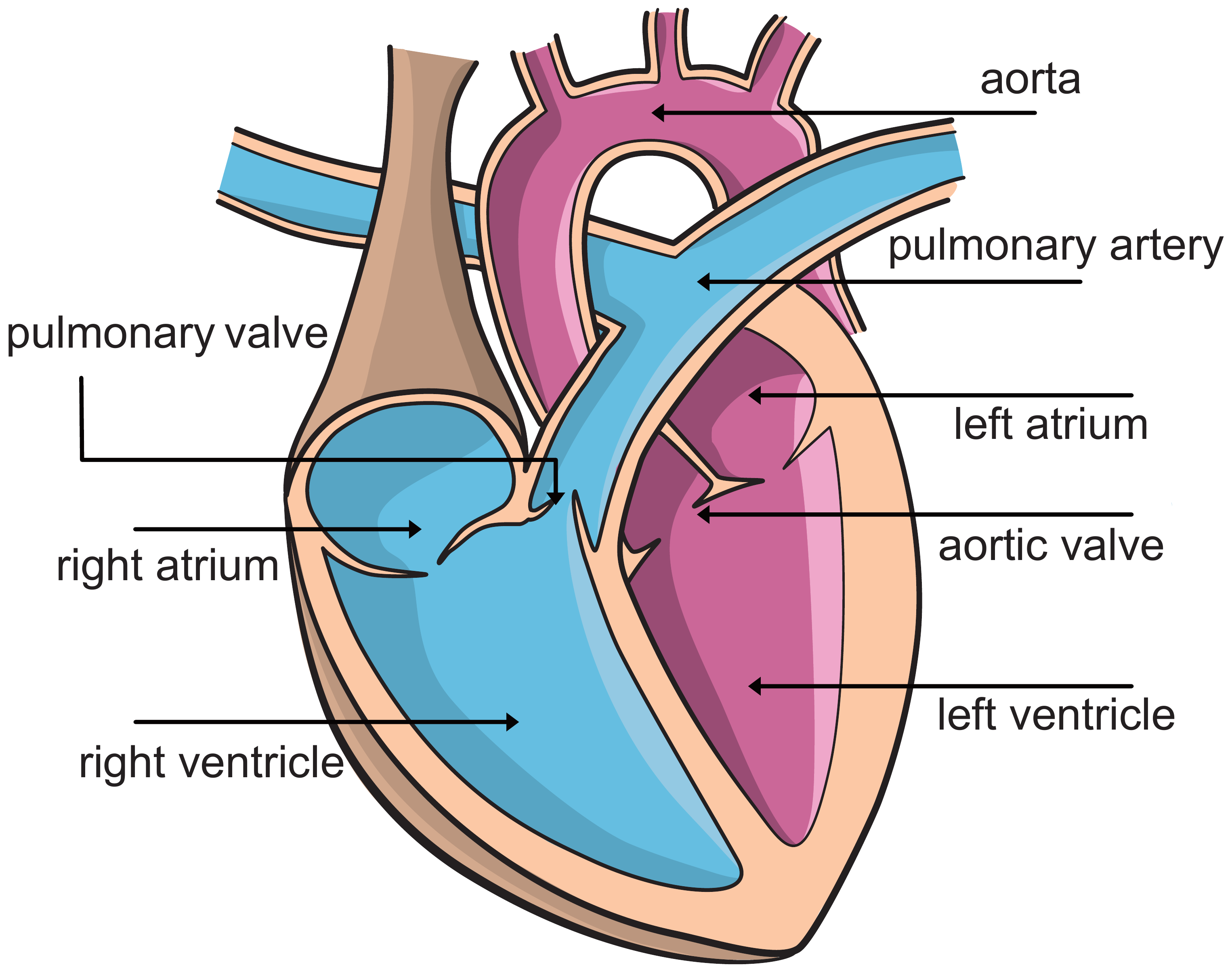} 
   \caption{
     \label{fig:heart}
     Anatomy of the human heart.
    }
\end{figure}

The vascular system forms an extensive network and, with the heart as the 
central pump, ensures the supply of all organs and tissues with oxygen and vital 
nutrients. 
The aorta is the largest artery in this system.
Their diameter of about 2-3 cm allows the acquisition of a measurable signal for 
a meaningful blood flow analysis.
During systole, oxygenated blood passes from the left ventricle (LV) via the aortic 
valve (AV) into the ascending aorta and is then supplied to the body, see Figure~\ref{fig:heart}.
From the right ventricle, deoxygenated blood is pumped through the pulmonary 
valve into the pulmonary artery.
During diastole, the AV is closed to prevent blood from flowing back into the LV. 

\begin{figure*}[t]
  \makebox[\textwidth]{\makebox[1.3\textwidth]{
    \includegraphics[width=1.3\linewidth]{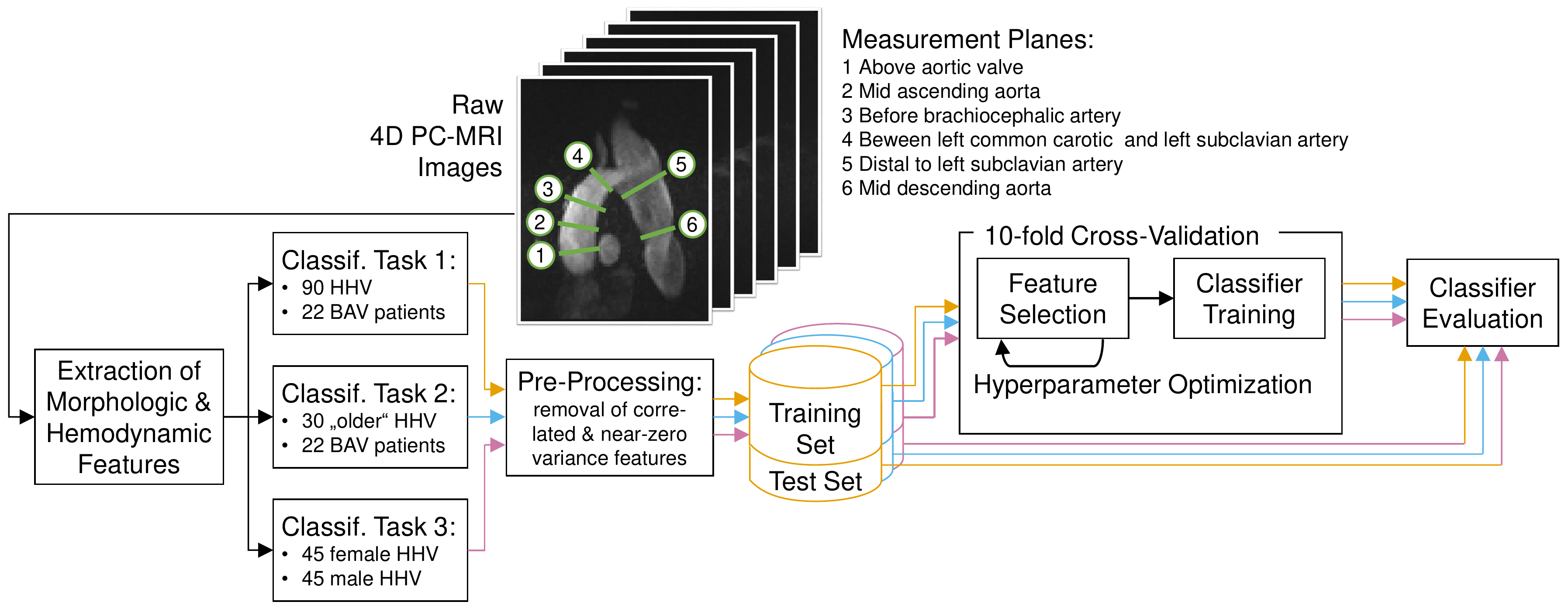}
  }}
   \caption{
     \label{fig:pipeline}
     \textbf{Schematic overview of our data analysis pipeline.} 
     For each of the 112 4D PC-MRI data sets, we extract morphological and hemodynamic features (197 in total). 
     We build separate classification models on three different subsets. 
     Therefore, we initially split the data into a training set (70\%) and a test set (30\%). 
     Then, we perform feature selection and classifier training. 
     Feature selection hyperparameters are tuned within a 10-fold cross-validation setup. 
     We use a wide range of combinations of feature selection methods and classification algorithms. 
     Finally, we identify the best model that achieves highest performance on the test set.
    }
\end{figure*}

\subsection{BAV and Related Flow Behavior}
\label{subsec:aorticpath}
Hemodynamic characteristics are an indicator for the emergence and progression 
of CVDs. 
Healthy people have only a slight systolic helix in the aortic arch, which 
is considered physiological.
CVDs change the vessel morphology, which increases the probability 
of quantitative and qualitative flow abnormalities~\cite{Kilner1998}.

The bicuspid aortic valve is the most common congenital heart defect with a 
prevalence of 1-2\%~\cite{Ward2000}. 
Normally, an aortic valve consists of three leaflets, whereas in a BAV two of them 
are fused together~\cite{Siu2010}. 
Due to the faulty morphology of the aortic valve, 
secondary diseases such as \textit{aortic valve stenosis} or 
\textit{aortic valve insufficiency} often develop over time. 
In aortic valve stenosis, a narrowing of the valve area occurs, so that the 
aortic valve can no longer open completely. 
In the case of aortic valve insufficiency, the aortic valve cannot close properly. 
As a result, blood in the diastole flows from the aorta back into the left 
ventricle - with the consequence of a volume strain on the left ventricle. 
 
Several studies investigated the influence of a BAV on flow patterns in the 
ascending aorta~\cite{Hope2010,Meierhofer2013,McNally.2017}. 
In comparison with healthy volunteers, patients more often had a helical 
blood flow, often accompanied by an eccentric main blood flow and increased 
wall shear stress. 
This results in permanent hemodynamic stress, which can lead to aortic valve 
stenosis, aneurysms, and/or AV insufficiency~\cite{Dux2019}. 
Recently, Ebel et al.~\cite{Ebel2020} performed a statistical analysis of 
different flow attributes between healthy and BAV patients, using standard 
techniques such as box plots. 
The greatest differences between the two groups were found in the flow velocities 
and the vortex volume.
However, the cohorts were quite small and there was no computer support for 
attribute selection, which made the analysis process time-consuming and error-prone. 
Thus, a detailed analysis of the flow attributes regarding several cohorts 
remains a challenge. 

\subsection{Data Acquisition and Feature Extraction}
A 4D PC-MRI data set contains each three (x-, y- and z-direction)
time-resolved phase and magnitude images that describe the flow direction
and strength, respectively.
All temporal positions together represent one full heartbeat.
A 3 T Siemens Magnetom Verio MR scanner was used with a maximum expected 
velocity ($\text{V}_\text{ENC}$) of 1.5 m/s per dimension.
The spatio-temporal resolution is 1.77 $\times$ 1.77 $\times$ 3.5 mm$^3$ $\backslash$ 
50 ms with a 132 $\times$ 192 grid for each of the 19 to 35 slices and 18 to 33 
time steps.

To analyze individual 4D PC-MRI data sets the software \emph{Bloodline}~\cite{Koehler2019} 
is used, which provides a guided, easy-to-use workflow. 
During pre-processing, phase wraps (aliasing) and magnetic coil artifacts are 
automatically corrected. 
Subsequently, the Graph Cut algorithm is employed to segment individual vessels, 
such as the thoracic aorta or pulmonary artery. 
This results in binary segmentation masks, from which polygonal vessel surfaces 
are extracted using Marching Cubes, which are post-processed by smoothing and reduction. 
In addition, a centerline is determined for each vessel. 

After pre-processing, the complete flow is automatically calculated within the 
vessel based on flow information encoded in the phase images. 
On the one hand, a qualitative analysis is performed to explore patient-specific 
flow patterns such as vortices, which may be an indicator of pathological 
changes~\cite{Hope2010}. 
On the other hand, quantitative methods allow the evaluation of the heart function. 
For this purpose, 197 attributes, such as heart rate, or maximum vessel diameter, 
commonly used in clinical studies, are calculated automatically. 

For the hemodynamic features, a distinction is made between two large groups: 
features that affect the vortical flow and features that describe the laminar flow. 
The proportion of vortical flow was determined using the $\lambda_2$ criterion. 
In addition, planes were placed in the aorta along the centerline. 
For each plane, different properties of the vortical flow were measured over 
the time steps. 
The maximum, minimum, median and mean values per feature were then determined 
for the set of result values over all planes and time steps. 
Since these properties are based on the planes along the aorta, they are also 
called ``in-plane'' features. 
In addition, so-called ``through-plane'' features, which characterize 
the laminar flow, are automatically calculated~\cite{Koehler2019} at specific 
landmarks. 
Subtracting the vortical portion from the total flow, the laminar flow 
portion is obtained. 
Similar to the in-plane features, the through-plane features are determined at 
each plane over time and the maximum, minimum, median and mean value per feature 
is calculated.
Finally, all data collected with Bloodline can be exported, enriched with additional 
information about the patient (age, gender, weight, etc.) and recording modalities 
(MRI sequence, recording duration, etc.).
So far, 90 data sets of heart-healthy volunteers and 22 BAV patients have been 
evaluated with Bloodline. 
These 112 data sets serve as input for our classification pipeline. 
Currently, we focus on the analysis of aortic flow data, as we have 
only few patients with pathological changes in other vessels such as the 
pulmonary artery.
Aortic diseases occur more frequently in clinical routine and are therefore 
our main focus.
\section{Methods}
\label{sec:framework}
Our data analysis pipeline is depicted in Figure~\ref{fig:pipeline}. %
After extracting the morphologic and hemodynamic features from raw 4D PC-MRI images, we build  separate models for each of three \emph{classification tasks}, i.e., models that distinguish  (1) HHV vs. BAV patients, (2) ``older'' HHV vs. BAV patients and (3) female HHV vs. male HHV.  
For this purpose, we make use of a wide range of feature selection methods and classification algorithms. 
This section describes the main steps of the pipeline, including data prep-processing, feature selection, classification and  evaluation. 

\subsection{Data Pre-Processing}
\begin{figure}[htb]
   \centering
   \includegraphics[width=0.5\linewidth]{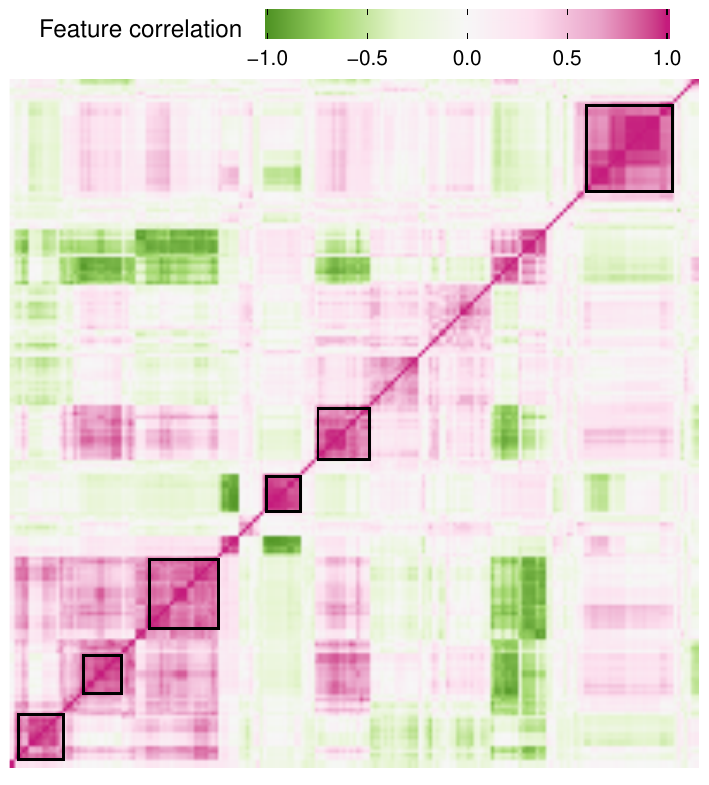}
   \caption{
     \label{fig:corplot}
     \textbf{Correlation heatmap.} 
     Spearman correlation heatmap for all 200 features (197 morphological and hemodynamic parameters, age, gender, and body weight). 
     Features are ordered by the result of agglomerative hierarchical clustering with complete linkage. 
     Subsets of at least 10 features with moderate to high pairwise correlation are enframed with black squares.
    }
\end{figure}

As part of data pre-processing, we remove highly correlated and low variance features, with the purpose of mitigating the negative effects caused by multicollinearity, and to prevent building unnecessary complex classification models. 
Ideally, our models achieve highly predictive performance and are as simple as possible at the same time, i.e., they should require only a small number of features. 
The correlation heatmap in Figure~\ref{fig:corplot} reveals 5 larger features subsets with moderate to high pairwise correlation. 
remove highly correlated features using the algorithm of Kuhn and Johnson~\cite{Kuhn2013}. 
For this purpose, we first compute Pearson correlation for each pair of features. 
If a pair of features exhibits an absolute correlation value of 0.9 or higher, we keep the feature that has the lower average correlation with the rest of the features. 
We repeat this process until none of the correlation values exceed the specified threshold.
In the process, we remove 103 of such features. 
Secondly, we eliminate near-constant features, where the most frequent value (rounded to two decimal places for continuous features) occurs more than 20 times as often as the second most frequent value and where the number of unique values is smaller than 10. 
We remove 3 such features, resulting in a total of 94 features for classification. 


\begin{figure}[htb]
   \centering
   \includegraphics[width=0.65\linewidth]{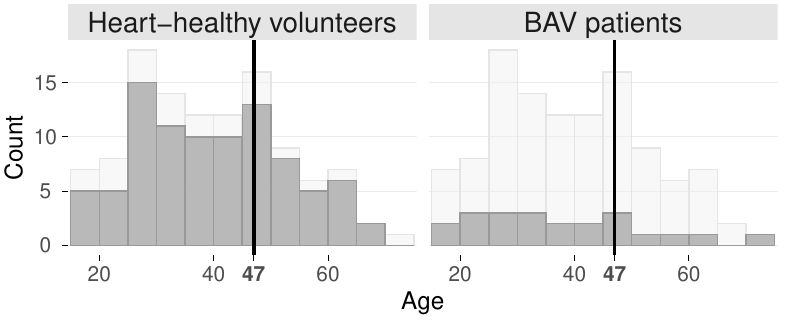}
   \caption{
     \label{fig:histogram-age-distribution}
     \textbf{Age distribution.} 
     According to information gain towards cardiac status, the optimal cutoff for a binary discretization is 47 years (right-closed).  
     We train a separate classification model that distinguishes between heart-healthy volunteers older than 47 years and all BAV patients. 
     Light gray bars show the overall age distribution.
    }
\end{figure}

For the second classification task, we divide the feature ``age'' into two intervals, which we refer to as ``younger'' and ``older'', respectively.
We determine the cutoff value as age with maximum information gain~\cite{Quinlan1986} towards cardiac status (cf. Figure~\ref{fig:histogram-age-distribution}).
Discretization leads to a subgroup of 78 younger subjects up to and including 
47 years (60 HHV, 18 BAV patients), and a subgroup of 34 older subjects (30 HHV, 4 BAV patients). 
The data was then divided into a training set and a testing set, with a ratio of 70:30 for further analysis.

\subsection{Feature Selection} 
We bundle classifier training with a preceding feature selection step that aims to reduce computational complexity, to increase model performance and enhance model interpretability. 
Feature selection methods can be broadly categorized into filter and wrapper methods~\cite{Jovic:FSReview2015,Kohavi:WrapperFS1997}.  
Filter methods rank features based on a specific importance measure, e.g., correlation with the target variable. 
The features that were found to be important are kept and then fed to the classification algorithm. 
Typically, the search is performed just once and independent from the classification algorithm. 
In contrast, wrapper methods use the performance of the classification model itself to measure the goodness of a candidate feature subset. 
Traditional wrapper methods work in an iterative, greedy fashion: the model is repeatedly trained on different feature subsets, where model performance on the current feature subset determines which subset to evaluate next. 
We use three filter and four wrapper methods.

\paragraph*{Correlation-based feature selection (CFS) filter} 
The non-parametric CFS~\cite{Hall:CFS2000} searches for a feature subset in which each individual feature is highly correlated towards the target variable but uncorrelated to every other feature within the subset. 
The ``merit'' of a feature is quantified as the quotient between feature-target correlation and feature-feature correlation. 
CFS starts with an empty set and iteratively adds the feature with maximum merit score until none of the remaining feature candidates has a positive merit score.  

\paragraph*{$\chi^2$-test filter}
In statistics, a $\chi^2$-test is used to test the independence of two categorical variables. 
This filter performs a $\chi^2$-test for every feature and the target feature. 
Features are ranked by p-value. 
Continuous features were discretized beforehand with the non-parametric, supervised method of Fayyad and Irani~\cite{Fayyad:MDL93}. 
The relative number of features that are retained for classification must be specified. 

\paragraph*{Information gain filter (IG)}
Similarly to the $\chi^2$-test, information gain measures the relationship between the target variable and a feature. 
It is calculated as reduction of uncertainty in the value of the target variable after observing the feature~\cite{Quinlan1986}. 
This filter calculates information gain for every feature with respect to the target variable. 
Just as the $\chi^2$-test, the percentage of selected features must be set.

\paragraph*{Sequential forward search (SFS) wrapper}
SFS is a greedy forward selection method that starts by training a classifier on an empty set of features. 
In each iteration, the feature that improves model performance the most is added to the model. 
If there is no more candidate feature whose addition leads to a relative increase of $\alpha$ in model performance, the method terminates.

\paragraph*{Sequential backward search (SBS) wrapper} 
SBS follows a search strategy that is reverse to SFS. 
It is a top-down approach which starts training a classifier on the whole feature set and iteratively removes the feature whose elimination leads to the least decrease in model performance.  
If there is no more candidate feature whose removal leads to an relative decrease of less than $\beta$ in model performance, the method terminates.

\paragraph*{Genetic search (GS) wrapper} 
GS methods are inspired by evolution processes found in nature. 
They emulate the mechanism of natural selection by iteratively performing so called crossover, mutation, elite selection and insertion operations on an initial set of candidate feature subsets~\cite{Oh:GeneticFS2004}. 
These operations involve adding and removing features from individual candidate subsets, creating new feature subsets by combining and swapping features from pairs or groups of candidate subsets, as well as removing some candidate subsets. 
A fitness function, usually the performance of the classification model, determines the probability of survival of a feature subset candidate for the next iteration. 
All operations include some aspects of randomization to avoid a solution from being stuck at a locally optimal feature set.  
The number of initial candidate feature subsets (``parent population'') and the maximum number of iterations were determined via hyperparameter tuning; 
crossover rate is set to 0.5 and mutation rate is set to 0.05.


\paragraph*{Random search (RS) wrapper} 
At each iteration, RS creates a feature subset candidate by random sampling from the total feature set. 
The model performance of a candidate set has no influence on the construction of the feature subset candidate of the next iteration. 
The number of iterations is varied and the probability of a feature to be randomly selected is 50\%. 
\subsection{Classification Algorithms} 
We utilized the following five classification algorithms: 
CART decision tree (DT)~\cite{BreimanEtAl:CART1984}, 
random forest (RF)~\cite{Breiman:RandomForests2001}, 
gradient boosted trees (GBT)~\cite{Friedman:PDP2001},
support vector machine (SVM)~\cite{Boser:SVM1992}, and 
least absolute shrinkage and selection operator (LASSO)~\cite{Friedman:LASSO2010}.

\paragraph*{Decision tree (DT)} A DT classifier recursively divides the data space into a set of non-overlapping subsets based on combinations of feature-value conditions, such as ``IF heart rate $>$ 95 bpm AND max vessel diameter $>$ 35 mm''.
These conditions are constructed in a way that the homogeneity of the target variable within the subsets is reduced. 
Each subset is assigned a class label which is usually the majority class of the training instances within that subset. 
\paragraph*{Tree ensembles}
Random forest (RF) and gradient boosted trees (GBT) classifiers are ensembles of multiple decision trees.
Each base tree casts a vote towards the prediction of the ensemble.  
In a RF base trees are constructed independently from each other, 
whereas trees in a GBT classifier are sequentially added to the ensemble, each of them focusing on areas of the data space where the previously trained trees exhibit high error.

\paragraph*{Support vector machine (SVM)}
In contrast to tree-based models, SVMs also capture non-linear relationships between features and target variable. 
For this purpose, they use a non-linear mapping to expand the feature space of the training set into a higher dimension. 
Within this new feature space, the optimal linear separating hyperplane, i.e., the decision boundary separating the observations from different classes, is detected. 

\paragraph*{Least absolute shrinkage and selection operator (LASSO) classifier}
LASSO is an extension of ordinary least squares linear regression (OLS) whose target function contains an additional regularization term that aims to shrink model coefficients of irrelevant features towards 0. 
Due to its intrinsic feature selection mechanism, LASSO models often are both simpler and more accurate in comparison to OLS~\cite{Hastie:ESL2009}.

\subsection{Evaluation}
We train separate models for each combination of feature selection method (7) and classification algorithm (5), yielding a total of 35 models. 
Hyperparameter selection for the feature selection methods were performed using 10-fold stratified cross-validation. %
We use accuracy as primary evaluation measure for the third classification task which exhibits a balanced class distribution (45 female HHV vs. 45 male HHV). 
Accuracy quantifies the ratio of correctly labeled observations to the total number of observations. 
Due to the class imbalance for the first two classification tasks (90 HHV vs. 22 BAV patients; 30 older HHV vs. 22 BAV patients), we opt for Cohen's kappa, respectively. 
Cohen's kappa is defined as:  
\begin{align}\label{Eq:ScalarVector}
\kappa = \frac {p_{o}-p_{e}}{1-p_{e}}, 
\end{align}
where $p_{o}$ represents accuracy and $p_{e}$ represents the probability for agreement among the vector of true class labels and the class labels assigned by random prediction.
In addition to accuracy and Cohen's kappa, we report area under the receiver operating characteristic curve (AUC). 
A receiver operating characteristic curve visualizes true positive rate (TPR) and false positive rate (FPR) for different prediction thresholds of a binary classifier. 
The area under the ROC curve (AUC) lies between 0 (0\,\% TPR, 100\,\% FPR) and 1 (100\,\% TPR, 0\,\% FPR), where a classifier that makes random predictions achieves an AUC of 0.5.

\section{Results}
\label{sec:experiments}
We summarize our classification results in Table~\ref{tab:performance}. 
Table~\ref{tab:hp-tuning} shows the feature selection hyperparameters that were tuned via cross-validation.

\paragraph*{Classification task 1: heart-healthy volunteers (HHV) vs. BAV patients} 
The best model (SFS + RF) achieves Kappa = 0.969 (accuracy = 93.9\%, AUC = 0.796) 
on the test set, cf. Table~\ref{tab:performance}. 
Three features were selected:
\begin{itemize}
\item \textsf{Time-to-Peak-Vorticity}: represents the time point during the heart 
cycle where the overall volume of blood swirling within a vortex in the aorta reaches its peak.
(in ms; 0 ms = begin of heart cycle).
\item \textsf{Time-to-Peak-In-Plane-Velocity}: depicts the time point 
during the heart cycle where the maximum velocity of the in-plane blood 
flow occurs in the aorta (in ms). 
\item \textsf{Peak-Systolic-In-Plane-Mean-Velocity}: quantifies the highest of the mean in-plane blood flow velocities over all planes and systolic timesteps (in m/s).
\end{itemize}

\begin{figure*}[htb]
   \centering
    \makebox[\textwidth]{\makebox[1.4\textwidth]{
       \begin{minipage}{0.7\linewidth}
       \includegraphics[width=\linewidth]{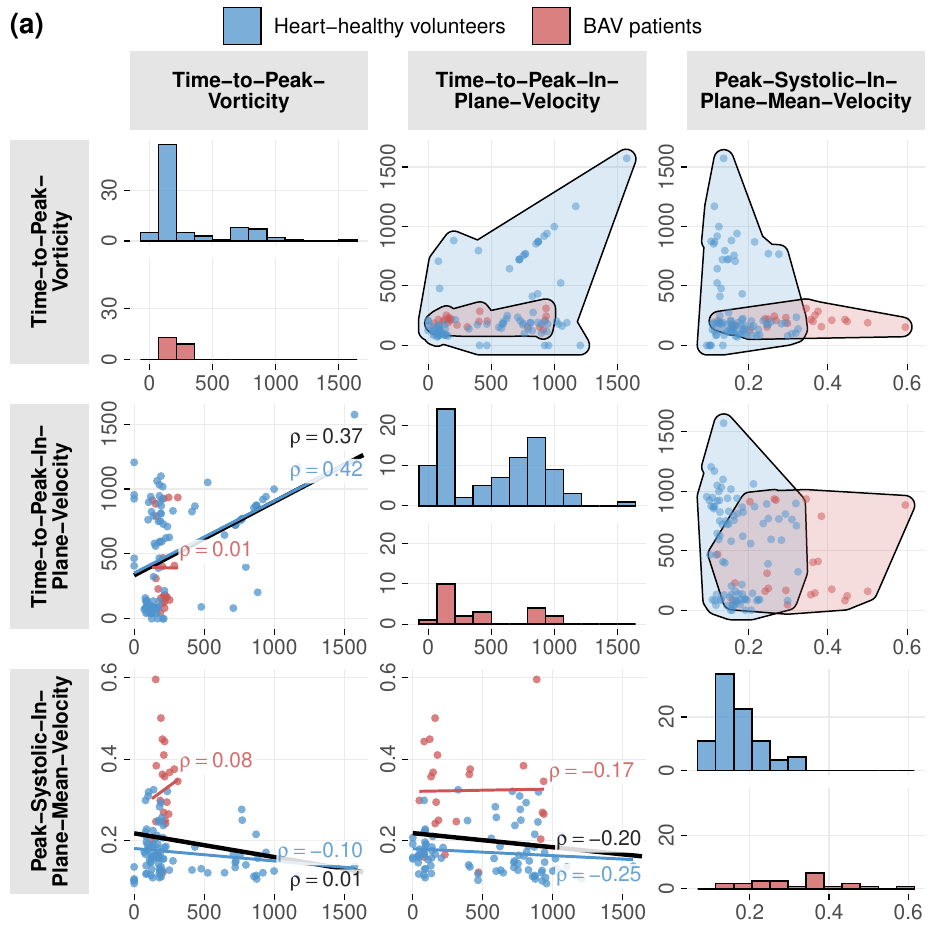}
       \end{minipage}
       \begin{minipage}{0.7\linewidth}
       \includegraphics[width=\linewidth]{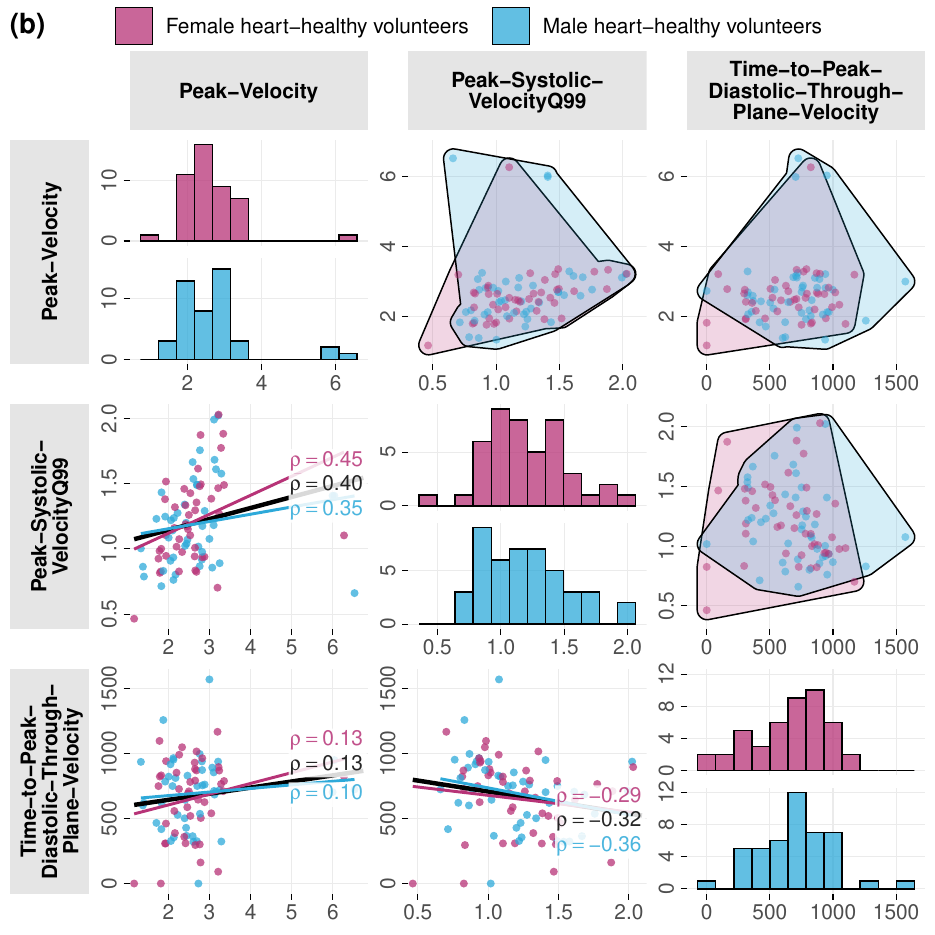}
       \end{minipage}
    }}
   
   \caption{
     \label{fig:plom-pathology-gender}
     \textbf{Plot matrix for features of best model for classification task 1 (SFS+RF) and 3 (SFS+SVM).}
     Pairwise relationships of the features that were selected for the best feature selection / classification algorithm combination for (a) classification task 1 and (b) classification task 3 are shown. 
     A scatterplot in the lower triangle of the matrix shows the relationship between the feature labeled at the top of the column (x-axis) and the feature labeled at the left of the row (y-axis).  
     Colored lines illustrate the strength of the linear relationship, and labels display the Spearman correlation coefficient ($\rho$). 
     Convex hulls in the scatterplots in the upper triangle of the matrix illustrate the overlap between the two subgroups in the depicted two-dimensional feature space.  
     Histograms on the main diagonal show the distribution of a feature for each target variable. 
    }
\end{figure*}

Figure~\ref{fig:plom-pathology-gender}~(a) highlights considerable differences 
between HHV and BAV patients with respect to these three features. 
In particular, BAV patients exhibit a much smaller range in \textsf{Time-to-Peak-Vorticity} 
(min: 130 ms, max: 312 ms) than HHV (min: 0 ms, max: 1573 ms). 
Similarly, whereas all HHV show a \textsf{Peak-Systolic-In-plane-Mean-Velocity} 
of 0.32 m/s or lower, variability within the BAV subgroup is much larger 
(min: 0.12 m/s, max: 0.60 m/s). 

\begin{table}[htb]
\caption{\label{tab:performance} \textbf{Classification results.} 
For each classification task, number of features $d$ and test performance of the five best combinations of feature selection method (FS) and classification algorithm (CA) are depicted. ACC: accuracy.
}
\centering
\begin{tabular}{c@{\hskip 0.20cm}r@{\hskip 0.20cm}l@{\hskip 0.20cm}r@{\hskip 0.20cm}r@{\hskip 0.20cm}r@{\hskip 0.20cm}r}
  \toprule
  \textbf{Classif. task} & \# & \textbf{FS + CA} & $\mathbf{d}$ & \textbf{Kappa} & \textbf{ACC} & \textbf{AUC} \\ 
  \midrule
  \multirowcell{5}{-1-\\HHV vs.\\BAV patients} & 1 & SFS + RF & 3 & \textbf{0.969} & 0.939 & 0.796\\ 
   & 2 & SFS + GBM & 3 & 0.962 & 0.939 & 0.819\\ 
   & 3 & SFS + SVM & 3 & 0.901 & 0.969 & 0.891\\ 
   & 4 & SFS + DT & 2 & 0.639 & 0.878 & 0.935\\ 
   & 5 & CFS + GBM & 19 & 0.796 & 0.939 & 0.898\\
   \midrule
   \multirowcell{5}{-2-\\``older'' HHV vs.\\BAV patients} & 1 & SFS + SVM & 5 & \textbf{1.000} & 1.000 & 1.000\\ 
   & 2 & SFS + RF & 3 & 0.925 & 0.866 & 0.722\\
   & 3 & SFS + DT & 1 & 0.777 & 0.800 & 0.571\\ 
   & 4 & SFS + GBM & 1 & 0.777 & 0.800 & 0.571\\ 
   & 5 & CFS + DT & 6 & 0.705 & 0.866 & 0.833\\
   \midrule
   \multirowcell{5}{-3-\\female HHV vs.\\male HHV} & 1 & SFS + SVM & 3 & 0.384 & \textbf{0.692} & 0.562\\ 
   & 1 & SFS + GBM & 3 & 0.384 & \textbf{0.692} & 0.562\\ 
   & 3 & SFS + RF & 2 & -0.076 & \textbf{0.692} & 0.384\\ 
   & 4 & CFS + GBM & 3 & 0.153 & 0.576 & 0.577\\
   & 5 & SFS + DT & 2 & 0.076 & 0.538 & 0.553\\ 
  \bottomrule
\end{tabular}
\end{table}

\renewcommand{\arraystretch}{1.1}
\begin{table}[htb]
\caption{\label{tab:hp-tuning} 
\textbf{Tuning grid of feature selection (FS) hyperparameters.} 
Values that yield best performance are highlighted in boldface.
}
\centering
\begin{tabular}{lp{7cm}l}
  \toprule
  \textbf{FS} & \textbf{Hyperparameter} & \textbf{Candidate values}\\ 
  \midrule
  $\chi^2$ & Relative number of selected features & \{0, $\frac{1}{9}$, $\frac{2}{9}$, $\mathbf{\frac{3}{9}}$, $\ldots$, 1\}\\
  IG & Relative number of selected features & \{0, $\frac{1}{9}$, $\frac{2}{9}$, $\mathbf{\frac{3}{9}}$, $\ldots$, 1\}\\  
  SFS & Minimum improvement in performance when adding a feature $\alpha$ & \{0.005, 0.01, \textbf{0.02}, 0.1\}\\  
  SBS & Maximum decrease in performance when removing a feature $\beta$ & \{$\mathbf{10^{-3}}, 10^{-2}, 10^{-1}$\}\\
  GS & Number of iterations & \{\textbf{10}, 20, 30\}\\  
  GS & Initial number of feature subsets & \{\textbf{10}, 15, 20\}\\  
  RS & Number of iterations & \{10, 50, 100, \textbf{200}\}\\  
  \bottomrule
\end{tabular}
\end{table}
\renewcommand{\arraystretch}{1}

\begin{figure*}[htb]
   \centering
   \makebox[\textwidth]{\makebox[1.3\textwidth]{
     \includegraphics[width=1.3\linewidth]{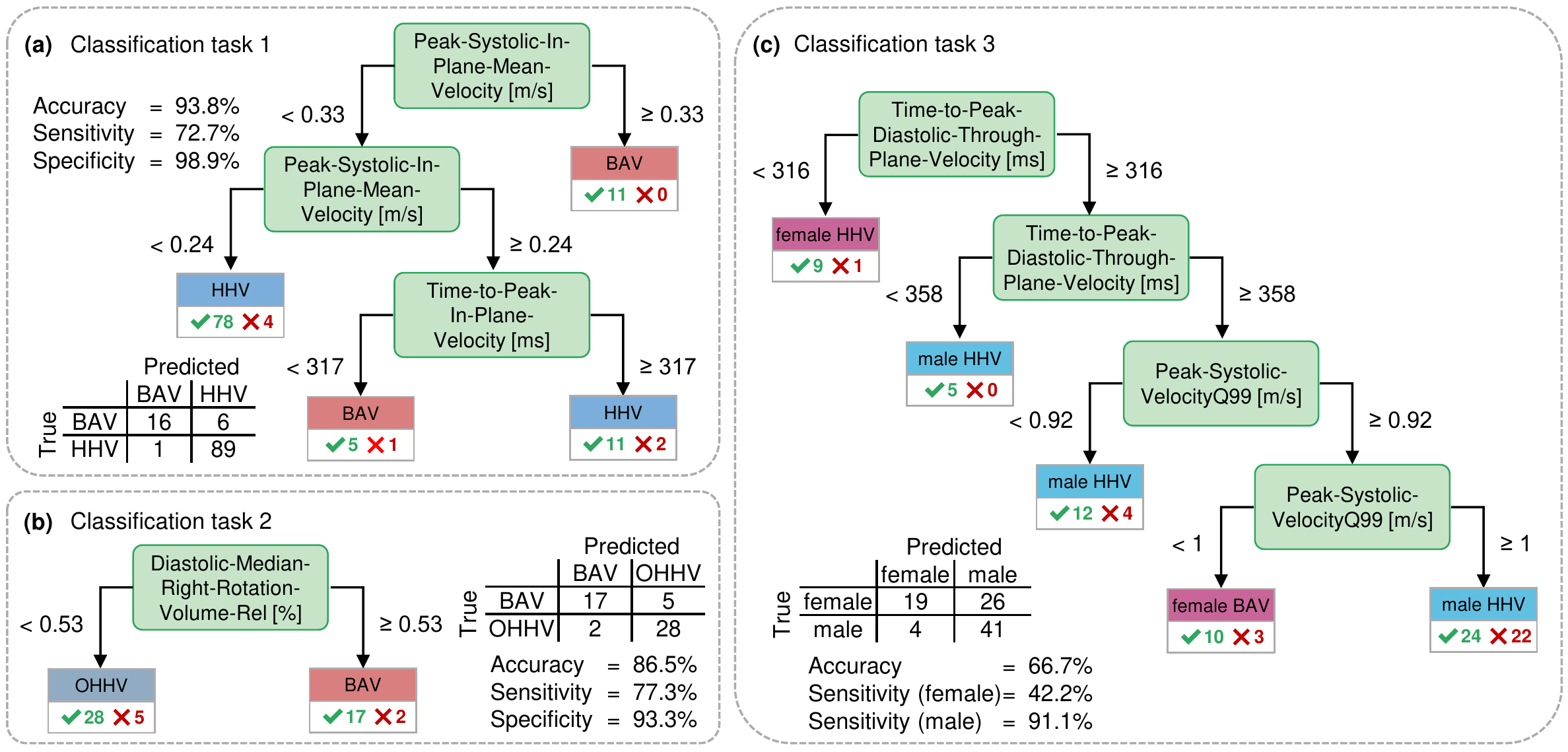}
   }}
   \caption{
     \label{fig:decision-trees}
    \textbf{Decision trees for each classification task.} 
    Simple decision trees were trained on the full data set using only the subset of features from the best combination of feature selection and classification algorithm. 
    Each terminal node depicts the predicted class label as well as the number of correctly and incorrectly labeled observations. 
    The minimum number of observations in a terminal node was set to 5; the maximum tree depth (number of levels) was set to 4. 
    OHHV: heart healthy volunteers older than 47.
    }
\end{figure*}

Visual inspection of the plot matrix in Figure~\ref{fig:plom-pathology-gender}~(a) 
suggests that a simpler model that uses the same feature set is capable to achieve 
high classification performance. 
Hence, for each classification task, we used the set of features that led to the 
best classification performance and constructed a decision tree 
104 out of 112 correctly classified instances). 
The tree in Figure~\ref{fig:decision-trees}~(a) shows descriptions of data regions 
which are representative for either of the two subgroups. 
For example, all 11 subjects with \textsf{Peak-Systolic-In-Plane-Mean-Velocity} 
$\geq$ 0.33 are BAV patients. 
In contrast, 78 out of 82 subjects exhibit a value  $<$ 0.24, cf. leftmost tree 
branch in Figure~\ref{fig:decision-trees}~(a). 
Overall, two features are sufficient to create a model with high accuracy (seven 
misclassifications out of 112 subjects).  

\paragraph*{Classification task 2: older HHV vs. BAV patients} 
The best model (SFS + SVM) correctly classifies all 15 test instances 
(Kappa = 1, accuracy = 100\%, AUC = 1). 
Five features were selected: 
\begin{itemize}
\item \textsf{Peak-Systolic-Mean-Velocity}: quantifies the highest of the mean through-plane blood flow velocities over all planes and systolic timesteps over a cardiac cycle in the aorta (in m/s). 
\item \textsf{Time-to-Peak-Systolic-Through-Plane-Mean-Velocity}: indicates the time point during 
the systole where the mean through-plane velocity over all planes' peaks (in ms). 
\item \textsf{Time-to-Peak-Diastolic-In-Plane-Mean-Velocity}: describes the time point during 
the diastole where the mean in-plane velocity over all planes' peaks (in ms).
\item \textsf{Diastolic-Median-Right-Rotation-Volume-Rel}: represents the median of 
the volume within the vessel that contains right-handed rotational flow during 
over all diastolic time steps in relation to the entire vessel volume (in \%). 
\item \textsf{Peak-Mean-Vorticity-Pressure}: measures the highest of the mean 
pressure value withing vortex regions over all time steps (in mmHg). 
\end{itemize}

\begin{figure*}[h!]
   \centering
   \includegraphics[width=\linewidth]{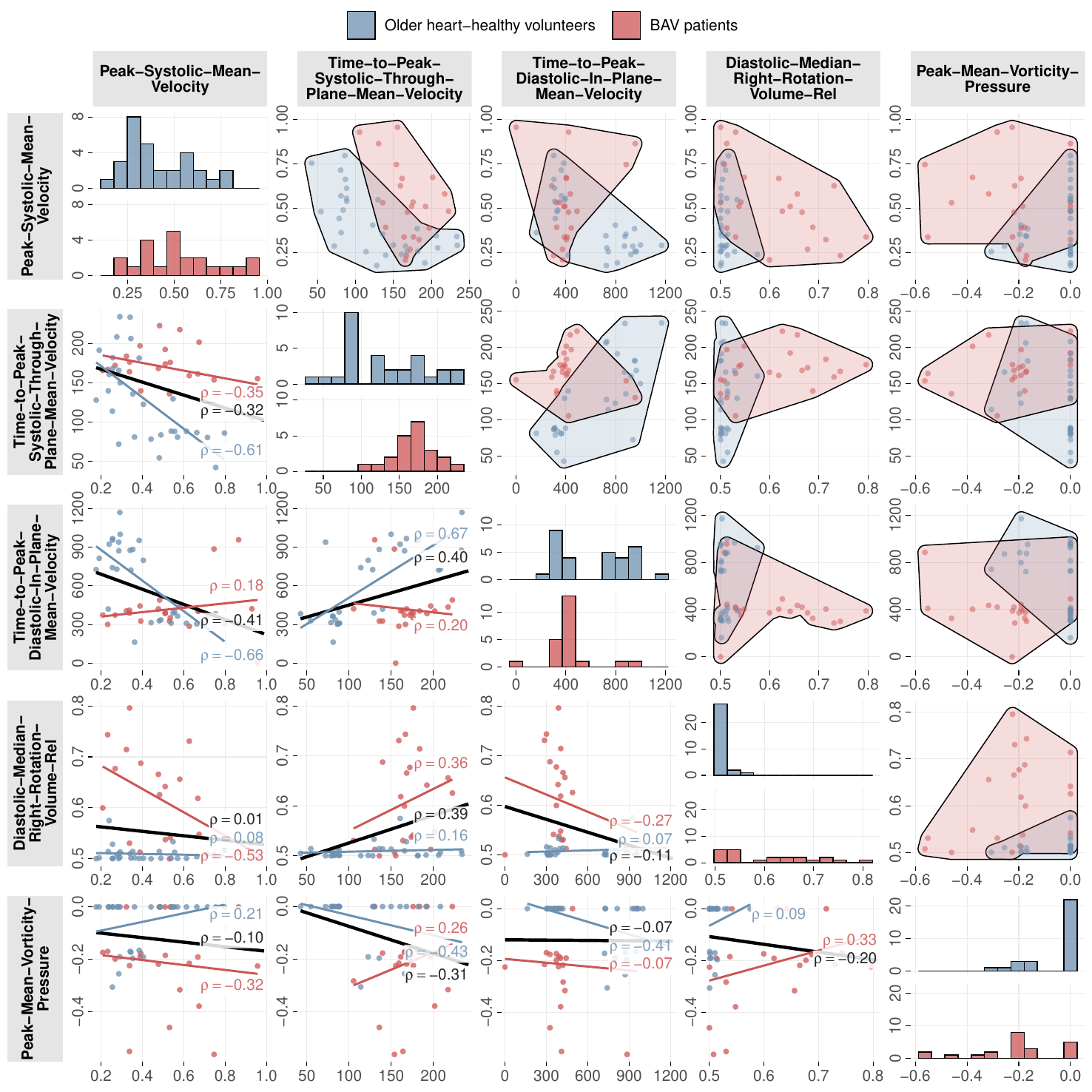}
   \caption{
     \label{fig:plom-ohhv_bav}
     \textbf{Plot matrix for features of best model of classification task 2 (SFS+SVM).}
     Pairwise relationships of the features that were selected for the best feature selection / classification algorithm combination of classification task 2 (sequential forward selection + support vector machine) are shown. See Figure~\ref{fig:plom-pathology-gender} for a description.
    }
\end{figure*}

Histograms and convex hulls in Figure~\ref{fig:plom-ohhv_bav} demonstrate large 
difference in the distribution of \textsf{diastolic-Median-Right-Rotation-Volume-Rel} 
between older HHV (mean: 0.51 $\pm$ standard deviation: 0.01) and BAV patients 
(0.61 $\pm$ 0.09). 
The first node of the decision tree depicted in Figure~\ref{fig:decision-trees}~(b), which is trained on the five above features, confirms the high association between \textsf{diastolic-Median-Right-Rotation-Volume-Rel} and subgroup membership. 
More specifically, a value greater equal than 53\% is characteristic for BAV 
patients (17 out of 22 subjects). 
Overall, the ``decision stump'' (only one split) in Figure~\ref{fig:decision-trees}~(b) 
correctly assigns 45 out of 52 subjects to their respective subgroup. 

\paragraph*{Classification task 3: female vs. male HHV}
The combination of SFS and RF achieves best generalization performance (accuracy = 69.2\%, Kappa = 0.384, AUC = 0.562), cf. Table~\ref{tab:performance}. 
Just as with the first classification task, three features were selected: 
\begin{itemize}
\item \textsf{Peak-Velocity}: measures the maximum velocity of blood flow 
that occurs in the aorta over a cardiac cycle (in m/s). 
\item \textsf{Peak-Systolic-VelocityQ99} depicts the maximum velocity of 
blood flow (taken from the 99.5\% quantile range to discard outliers due to noise) that occurs in the aorta during the 
systolic phase of the cardiac cycle (in m/s). 
\item \textsf{Time-to-Peak-Diastolic-Through-Plane-Velocity} represents the time point 
during the diastole where velocity of the through-plane blood flow peaks in any of the planes in 
the aorta (in ms; 0 ms = begin of heart cycle). 
\end{itemize}

Although these three features yield the best model, the plot matrix in Figure~\ref{fig:plom-pathology-gender}~(b) shows that female and male HHV have quite similar value distributions. 
Consequently, the decision tree trained on \textsf{Peak-Velocity}, \textsf{Peak-Systolic-VelocityQ99} and \textsf{Time-to-Peak-Diastolic-Through-Plane-Velocity} (Figure~\ref{fig:decision-trees}~(c)) is more complex than the other trees and has only mediocre classification performance, as one third of the subjects are misclassified.

\section{Discussion}
\label{sec:disc}
Our classification results suggest substantial differences in flow characteristics 
of the aorta between HHV and bicuspid aortic valve (BAV) patients. 
For several of the features, BAV patients exhibit higher variation and higher 
value ranges. 
To proof the validity of our models, we discussed the results with a radiologist 
(co-author), specialized in cardiac imaging with 23 years of work experience. 
He also provided the data sets. 

The possible aortic valve stenosis due to the altered valve morphology 
in BAV patients means that the heart must exert more force to open the valve and 
pump the blood further because of this resistance. 
Consequently, the flow velocity increases significantly compared to healthy 
individuals, especially during systole. 
The strong increase in flow velocity favors the formation of vortical blood flow 
within the aorta. 
As a result, the forward movement of the flow along the centerline decreases and 
there is more flow that rotates within a plane. 
This is also reflected in the classification results for the first task. 

The parameter \textsf{Peak-Systolic-In-Plane-Mean-Velocity} represents the 
velocity of flow in a plane. 
In HHV it is less than 30 cm/min and the average flow velocity is 1.5 m/sec. 
Accordingly, the velocity of flow in a plane in BAV patients is at least one 
third of the average flow velocity, which is a very high value. 
These circumstances explain the good initial separation of HHV and BAV patients 
based on \textsf{Peak-Systolic-In-Plane-Mean-Velocity}. 
The second parameter \textsf{Time-to-Peak-In-Plane-Velocity} measures 
the time of maximum flow velocity in one plane. 
The radiologist assumes that this is the corresponding time for 
the \textsf{Peak-Systolic-In-Plane-Mean-Velocity}. 
Due to the high flow velocity in BAV patients, vortical flow occurs earlier 
during the cardiac cycle and therefore the \textsf{Time-to-Peak-In-Plane-Velocity} 
occurs earlier than in healthy volunteers. 
Nevertheless, the radiologist was positively surprised that 
only two parameters allow a good separation of both classes. 
He emphasized 
that determining the corresponding threshold values 
is of paramount importance 
to support the diagnosis of CVDs. 

With respect to our second research question, we could not confirm that older HHV show similar flow characteristics as BAV patients in the current data set.
The parameter \textsf{Diastolic-Median-Right-Rotation-Volume-Rel} allows a reliable 
separation of both groups with 7 out of 45 data sets being misclassified. 
In HHV, only during systole slight vortical flow occurs, whereas during diastole 
hardly any vortical flow occurs because the aortic valve is completely closed. 
This is also the case in older subjects. 
BAV patients on the other hand show increasing vortical flow, also during diastole, 
which is why a differentiation of both groups was possible here. 
Since the best models have excellent performance while requiring only three to 
five features, we conclude that our approach to extract interpretable features 
for the diagnosis of BAV patients is sensible. 

From the performance of the classifiers for the 
third classification task, we 
infer that, although there are minor differences between female and male HHV, 
morphological and hemodynamic characteristics alone are not sufficient to distinguish 
between genders with high accuracy. 
This also met the expectations of our radiologist. 
Although on average men have a larger heart and blood volume than women, this 
does not significantly affect the derived morphological and hemodynamic characteristics. 
Nevertheless, such a detailed analysis of HHV is important to derive 
normal values and thus reliably detect abnormal characteristics. 

Although we consider our results to be promising, there are some limitations and 
room for improvement. 
First, our data set might be subject to a selection bias. 
The HHV group includes only volunteers who are at least 18 and at most 65 years 
old. 
Older volunteers were excluded because they had already shown an enlargement 
of the aortic diameter which, although not yet in need of treatment, deviated 
from a physiological morphology. 
Accordingly, our calculated age limit of 47 years for classifying older HHV and 
BAV patients was rather low, which might explain the good separability. 
Regarding BAV cases only patients over 18 years were considered. 
Children were therefore not included in the data collective, although it would 
be an interesting question whether children could be separated as well as adult 
BAV patients from HHV. 
In addition, no distinction was made in the BAV patients regarding 
pronounced secondary diseases, aortic valve stenosis and aortic valve insufficiency. 
This could also influence the classification. 

We used the Information Gain measure to determine the cutoff value of 47 years to divide the subjects into two subsets based on age. 
Alternatively, we could have also tested multiple thresholds to ultimately choose the cutoff value which yields best classification performance.   
However, a study involving coronary artery disease patients~\cite{Kuecherer1988} also showed that patients of ca. 50 years of age and older had similar ventricular diastolic filling to the control group, while there were larger differences between groups in younger subjects.

While the decision trees in Figure~\ref{fig:decision-trees} illustrate the protectiveness of some of the extracted features towards the target variables, 
they should be interpreted with caution, as they were trained on the total data set, and thus, have limited generalization value.

In addition, our holdout evaluation might be subject to high variability of the 
generalization error estimates, especially for classification task 2 where the test 
set contains only 15 observations (ca. 30\%). 
To assess the performance gain of our feature selection step, we also consider to train classifiers using all available features. 
We did not attempt this here, as the size of the feature space would have been ca. twice as large as the size of the data set and the performance estimates would have been even more prone to the effects of overfitting.
In future, we would like to expand our data analysis pipeline with a nested-cross 
validation scheme that involves tuning of feature selection hyperparameters in 
the inner loops and calculation of generalization performance on multiple outer folds, 
thus reducing variability of the performance estimates. 
Apart from parameters of the feature selection methods, we will consider also 
tuning classifier hyperparameters to further improve model performance. 
\section{Conclusion and Future Work}
\label{sec:con}
We present a data analysis pipeline for the classification of BAV patients vs. HHV. 
Therefore, we initially extract a large number of morphological and hemodynamic 
features from 4D PC-MRI data sets. 
We integrated various feature selection methods with classifier training to 
reduce the number of features, aiming to build accurate yet parsimonious classification 
models. 
Finally, we constructed decision trees using the subset of features found to 
achieve best classification performance. 
Our results show differences between BAV patients and HHV with respect to 
hemodynamics, such as the higher blood flow velocity in BAV patients. 
Furthermore, we applied our pipeline to two other classification tasks. 
First, we investigated the question whether there are morphological and 
hemodynamic differences between \emph{``older''} HHV and BAV patients, which was 
derived from the observation that the heart function seems to decrease with age. 
We were able to confirm this based on the high performance of the classifiers, 
although our results must be verified on another independent data set. 
Second, we trained a model that distinguishes between female and male HHV based 
on the extracted features. 
Although our models are more competitive than random guessing, we did not identify 
one or a combination of features that are capable to predict gender very accurately. 

In the future, we plan to extend our pipeline to other pathologies, e.g., diseases 
of other heart vessels, such as the Tetralogy of Fallot, which primarily show 
changes in the pulmonary artery. 
This may contribute to improve the diagnostic workflow.
Furthermore, we want to expand our data collection to include more older 
heart-healthy volunteers (60 years of age and older) and repeat the second classification 
task to check whether healthy aging people show similar characteristics to BAV 
patients.
Moreover, we plan to evaluate the treatment success, e.g., after an aortic valve 
replacement, to check whether the flow normalizes after surgery. 
For this purpose, we want to apply our trained models to new treated cases and 
check if they are classified as heart-healthy. 
This would provide insights into whether the flow returns to normal after the 
insertion of an artificial heart valve.  

\section*{Funding}
This work was partially supported by the Carl Zeiss Foundation.



\end{document}